\documentclass[letterpaper, 10 pt, journal, twoside]{IEEEtran}  

\usepackage{graphicx}
\usepackage{verbatim}
\usepackage{array}
\usepackage{upgreek}
\usepackage{float}   
\usepackage{amsmath,amssymb}
\usepackage{cite}
\usepackage{booktabs}
\usepackage{algorithm}
\usepackage[noend]{algpseudocode}
\usepackage{stfloats}
\usepackage{tabularx}

\usepackage{todonotes}
\usepackage{xcolor}

\usepackage[normalem]{ulem}

\makeatletter
\def\algbackskip{\hskip-\ALG@thistlm}
\makeatother

\usepackage{tikz}
\usetikzlibrary{positioning, shapes, arrows.meta}
\usetikzlibrary{shapes,arrows,chains}
\usetikzlibrary{arrows,calc,automata}

\tikzset{
    block/.style = {draw, rectangle, 
        minimum height=.5cm, 
        minimum width=1cm},
    input/.style = {coordinate,node distance=.5cm},
    output/.style = {coordinate,node distance=.5cm},
    arrow/.style={draw, -latex,node distance=.5cm},
    pinstyle/.style = {pin edge={latex-, black,node distance=1cm}},
    sum/.style = {draw, circle, node distance=1cm},
    line/.style={-{Stealth}}
    }

\begin{document}

\title{
Design and Flight of an Ion-propelled Micro Hovercraft Leveraging Ground Proximity Effects
}

\author{C. Luke Nelson$^{1}$, Grant Nations$^{1}$, Mrinmoy Modak$^{2}$, and Daniel S. Drew$^{2}$%
\thanks{Manuscript received: March 16, 2026; Revised June 7, 2026; Accepted June 28, 2026.}
\thanks{This paper was recommended for publication by Editor Xinyu Liu upon evaluation of the Associate Editor and Reviewers' comments.} 
\thanks{$^{1}$C. L. Nelson and G. Nations are with the Robotics Center, University of Utah, Salt Lake City, UT 84112 USA}
\thanks{$^{2} $M. Modak and D. S. Drew are with the Department of Electrical and Computer Engineering, University of Hawaii at Manoa, Honolulu, HI 96822 USA
        {\tt\footnotesize ddrew@hawaii.edu}}%
\thanks{Digital Object Identifier (DOI): see top of this page.}
}
%
%

\markboth{IEEE Robotics and Automation Letters. Preprint Version. Accepted June 2026}
{Nelson \MakeLowercase{\textit{et al.}}: Ion-propelled Micro Hovercraft} 

\maketitle

\begin{abstract}
Electroaerodynamic propulsion is compelling for use in micro air vehicles due to its silent and solid-state nature, but its limited efficiency has thus far precluded a path towards power-autonomous flight. Recent work has shown that thrust density and efficiency for small-scale atmospheric ion thrusters can be vastly increased when operating close to a ground plane. Here, we explore the design space of centimeter-scale hovercraft, which can leverage this ground effect for low-altitude flight. We first perform an empirical investigation, characterizing the performance benefits and trade-offs for different geometries and configurations of passive hovercraft skirts, then use the results to fabricate a viable point design. We demonstrate a palm-sized hovercraft that, while tethered to an external power source,  can fly for extended periods, withstand dozens of takeoff and landing cycles, passively stabilize to reject significant mechanical disturbances, and generate practically zero audible noise signature. The measured thrust efficiency of 16 mN/W and additional payload capacity of almost 1.5 grams above the vehicle's self mass of about 1.6 grams exceeds any similarly sized electroaerodynamically propelled robot by an order of magnitude. This is the first time an ion-propelled micro hovercraft has been shown in the open literature, and our work points the way towards an entirely new class of robot. 

\end{abstract}



\section{Introduction}
\IEEEPARstart{L}{arge} numbers of small robots are increasingly being used in close proximity to humans and the built environment. At the micro air vehicle (MAV) scale, rotorcraft (“drones”) dominate the current commercial and research landscape~\cite{floreano_science_2015,kumar_opportunities_2012}. Two significant challenges with the use of drones in constrained multi-occupant spaces (e.g., a warehouse with human workers) are the fragility of their quickly spinning propellers—one collision can mean catastrophic failure—and the high level of stress and annoyance caused by their rotor noise~\cite{schaffer_drone_2021}. The latter, especially, may prove an insurmountable barrier; a future with ubiquitous human-robot teaming in the workplace is only possible if robots can be tolerated. A viable alternative propulsion technology is needed.

The electroaerodynamic (EAD) force produced by momentum-transferring collisions between electrostatically accelerated ions and neutral air molecules has been harnessed for flying vehicles of all sizes, from centimeter-scale “ionocraft”~\cite{drew2018toward} to meter-scale “solid-state aircraft"~\cite{xu2018flight}. Atmospheric ion actuators are interesting because they are silent, solid-state, and scalable. Small multi-stage actuators have been shown with thrust densities approaching or exceeding state-of-the-art flapping wings and rotors~\cite{nelson2024high}. Still, limited thrust efficiency remains a significant barrier, and there is currently no feasible path towards a power autonomous vertical takeoff and landing (VTOL) ion-propelled robot at the MAV scale~\cite{grosse_modeling_2024}. Although larger fixed-wing vehicles may be viable using multi-stage ducted (MSD) actuators~\cite{gomez-vega_order--magnitude_2024}, they are not suitable for slow indoor use.

Surface proximity effects arising from aerodynamic interactions between vehicle airfoils and airframes with fixed obstacles can significantly alter flight performance~\cite{powers_influence_2013,matus-vargas_ground_2021}. Exploiting beneficial effects of surface proximity has been shown to improve the maneuverability, flight duration, and task performance of rotorcraft in constrained environments~\cite{gao2019exploiting}. New types of MAV-scale rotorcraft have been created specifically to make more effective and safe use of these proximity effects~\cite{hsiao2018ceiling,ding_passive_2022,hsiao_energy_2023}. These vehicle designs may reduce the incidence of propeller collisions, but do not solve the problem of noise. Flapping wing micro air vehicles (FMAVs) may also benefit from certain surface proximity effects while generating substantially less noise than rotorcraft~\cite{truong_aerodynamic_2013}, and soft dielectric elastomer-based designs have been shown to be resilient to collisions~\cite{ren2022high}. Nevertheless, the mechanical complexity of FMAVs, together with their need for high-speed control to maintain stability, may prove significant barriers to widespread use in the near future.

\begin{figure}[t]
    \centering
    \includegraphics[width=\columnwidth]{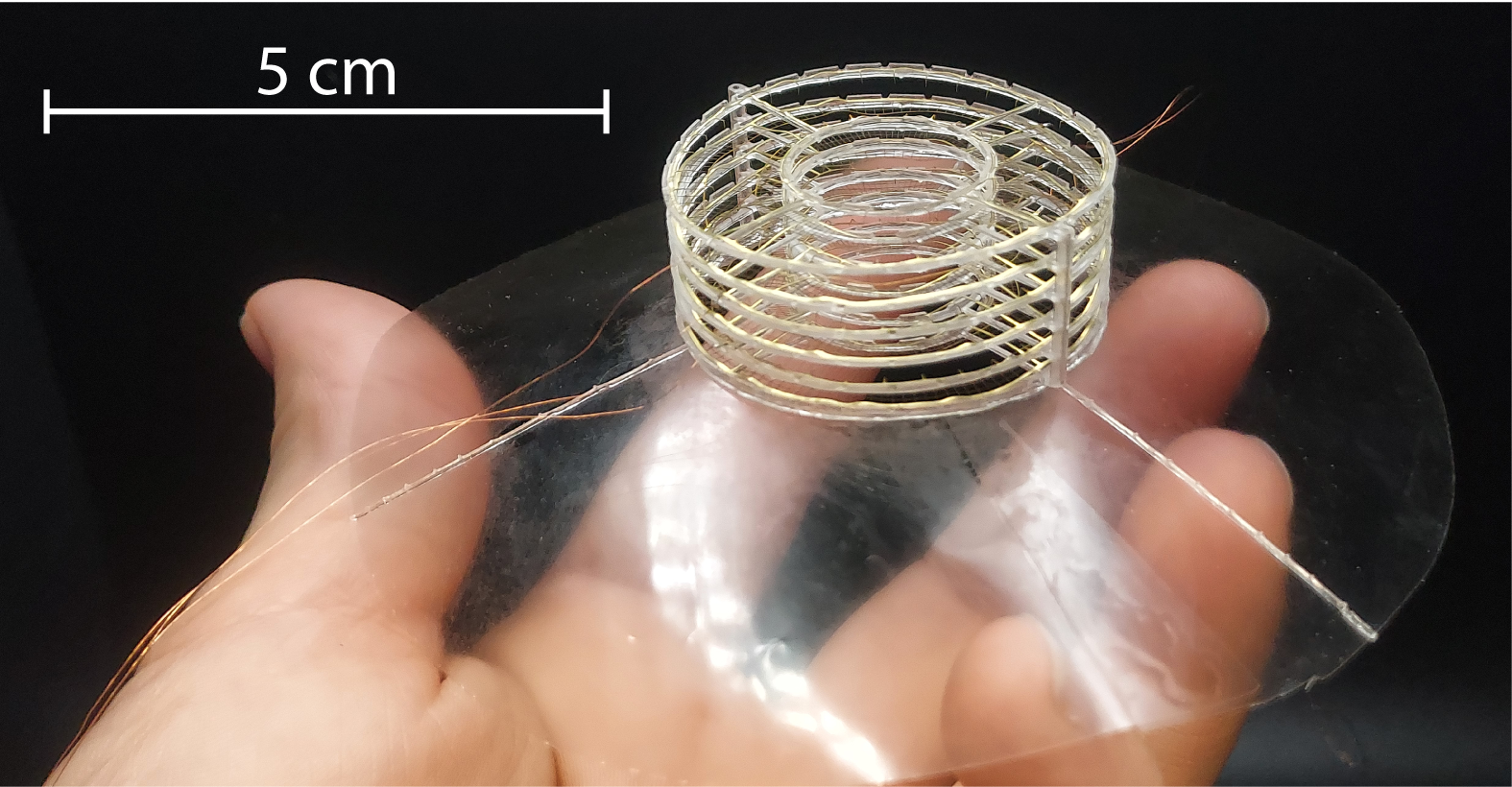}
    \vspace{-5.5mm}
    \caption{A palm-sized ion-propelled micro hovercraft capable of sustained tethered flight and passive disturbance rejection without closed-loop control.}
    \label{fig:teaser}
    \vspace{-5.5mm}
\end{figure}

Recent work showed for the first time that electroaerodynamic actuators, when properly designed, experience levels of beneficial ground proximity effects unmatched by either rotorcraft or FMAVs~\cite{nations2024empirical}. The resulting boost in performance could be what is required to make EAD a truly viable alternative propulsion scheme by raising efficiency to the levels needed for autonomy. This work served as the impetus for us to investigate the design of an ion-propelled hovercraft platform, with a central EAD thruster core and a passive skirt, to maximize ground effect benefits.

The primary contribution of this paper is the design and flight demonstration of a centimeter-scale electroaerodynamically propelled hovercraft that intentionally exploits ground proximity effects for lift. We empirically evaluate how skirt geometry and center-plate placement affect pressure-supported lift, then use these results to fabricate a free-standing vehicle. The resulting palm-sized platform achieves sustained tethered hover, repeated takeoff and landing, and passive disturbance rejection without closed-loop control. We further characterize its acoustic signature, flight dynamics, thrust efficiency, and payload capacity, showing performance well above prior similarly sized EAD-propelled fliers. Together, these results establish a path toward a new class of silent, solid-state micro air vehicle.

\begin{figure}[t]
    \centering
    \includegraphics[width=\columnwidth]{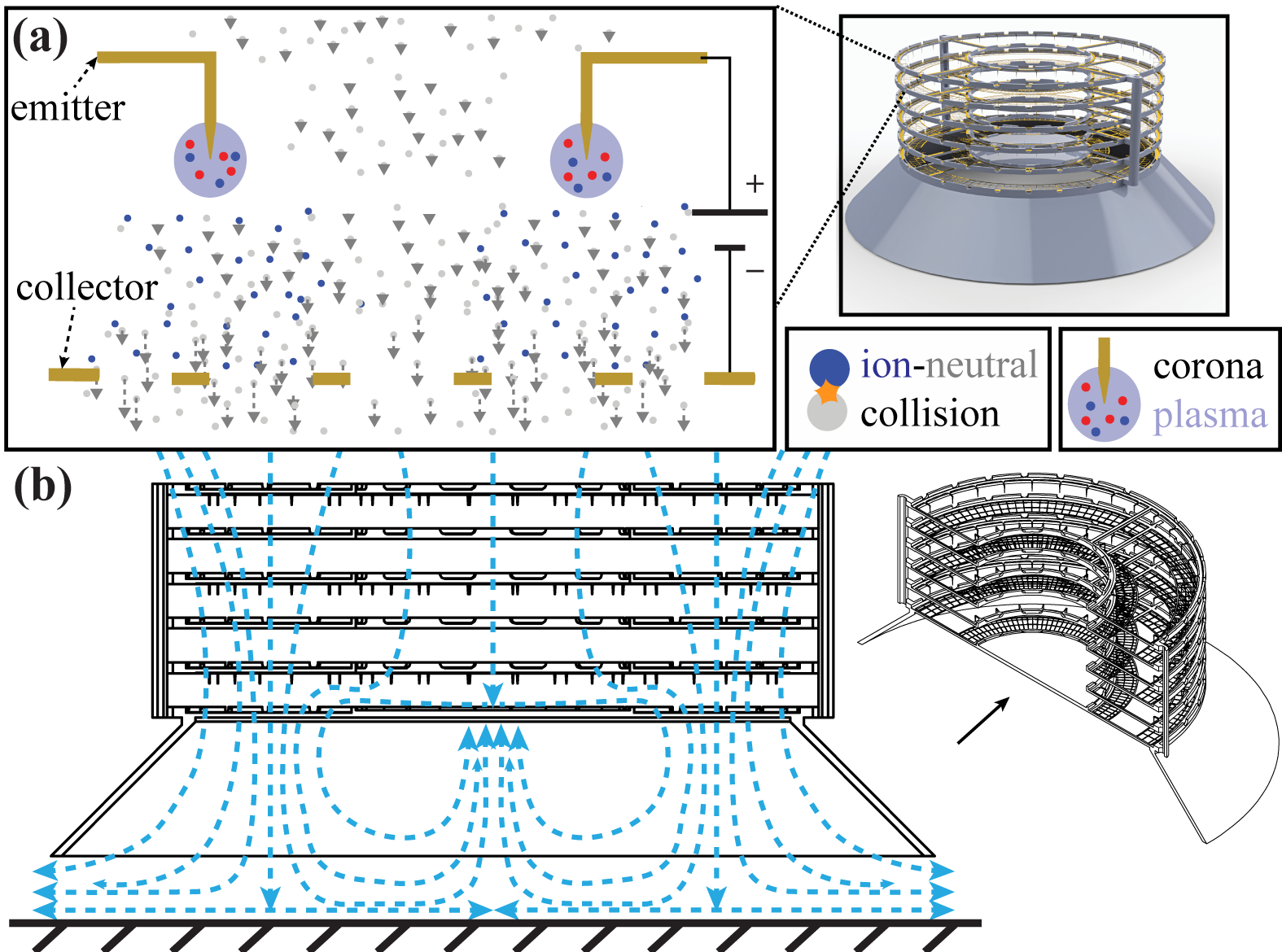}
    \vspace{-6mm}
    \caption{\textbf{(a)} Schematic of one of the three ion acceleration stages. Ions are ejected from the corona plasma around the emitter tips and collide with neutral air molecules, producing a net airflow through the collector grid.  \textbf{(b)} Depiction of the induced airflow through a cross-section of the hovercraft (inset). Air is accelerated through the thruster stages and impinges on the ground plane to form a wall jet. Wall jet collisions produce fountain lift.}
    \label{fig:schematic}
    \vspace{-5.5mm}
\end{figure}

\section{Background and Related Work}
\label{sec:background}

\subsection{Electroaerodynamic Propulsion}

Electroaerodynamic propulsion generates thrust through ion acceleration and momentum transfer to neutral air (Fig.~\ref{fig:schematic}(a)). Because they contain no moving components, EAD systems operate with low acoustic signature and structural simplicity. Early ``ionocraft'' demonstrations established centimeter-scale liftoff and controlled flight using corona-discharge actuators \cite{drew2018toward}, but thrust density and efficiency limited payload capacity and endurance.

Subsequent work improved performance through electrode optimization and multi-stage stacking \cite{nelson2024high}, increasing total momentum transfer while maintaining discharge stability. These architectures primarily target free-stream (out-of-ground-effect) operation, with proximity treated as a secondary effect. Although several MAV-scale EAD flying robots have been shown \cite{drew2018toward, nelson2024high,prasad_laser-microfabricated_2020,gu_microrobotic_2024,zhang_centimeter-scale_2022}, lift generation has relied on free-stream thrust, and performance remains constrained by the thrust–efficiency tradeoff inherent to space-charge-limited drift. To date, no MAV-scale EAD vehicle has used ground-mediated pressure recovery for flight.

\subsection{Proximity-Mediated Lift Enhancement}

Proximity to solid boundaries fundamentally alters aerodynamic performance. In rotorcraft, classical models relate thrust augmentation to reduced induced velocity near the surface~\cite{cheeseman_effect_1955}. Small-scale and multirotor experiments reveal additional phenomena, including fountain lift and recirculating wall jets, not captured by single-rotor theory~\cite{conyers_empirical_2018}. When downward flow is laterally confined by ducts, shrouds, or skirts, proximity effects can transition from induced-velocity reduction to pressure-supported lift. Impinging wall jets spread radially along the ground, generating elevated static pressure beneath the vehicle. Lift then depends on both thrust magnitude and the footprint area over which pressure acts. Traditional air-cushion vehicles exploit this mechanism directly by mechanically confining airflow beneath a skirt to form a pressurized cushion~\cite{ganesan_design}.

Prior hovercraft and air-cushion vehicle research has focused on dynamics, control, and skirt optimization, typically at scales much larger than MAVs~\cite{xie_robustcontrol,ganesan_design,chung_optimization}. Existing unmanned hovercraft prototypes employ fan-driven pressurization and conventional skirts~\cite{roubieu_fully-autonomous_2012}. While these studies confirm that geometric confinement converts downward momentum into pressure-supported lift, they do not address extremely size- and mass-constrained vehicles, or ones in which EAD actuators generate the flow.

\vspace{-2mm}
\subsection{Proximity Effects in Electroaerodynamic Systems}

Only recently have proximity effects been examined for EAD propulsion~\cite{nations2024empirical,nelson_empirical_2024}. Multi-stage thrusters were shown to exhibit thrust amplification in-ground-effect due to wall-jet interaction and pressure recovery which exceeds the values that have been measured for similarly sized rotors. These studies, however, characterized these effects only for isolated thrusters in fixed test rigs, and did not intentionally integrate proximity effects into a self-supporting vehicle architecture as in this work (Fig.~\ref{fig:schematic}(b)).

\section{Methods}
\label{sec:methods}

\begin{figure*}[t]
    \centering
    \includegraphics[width=\textwidth]{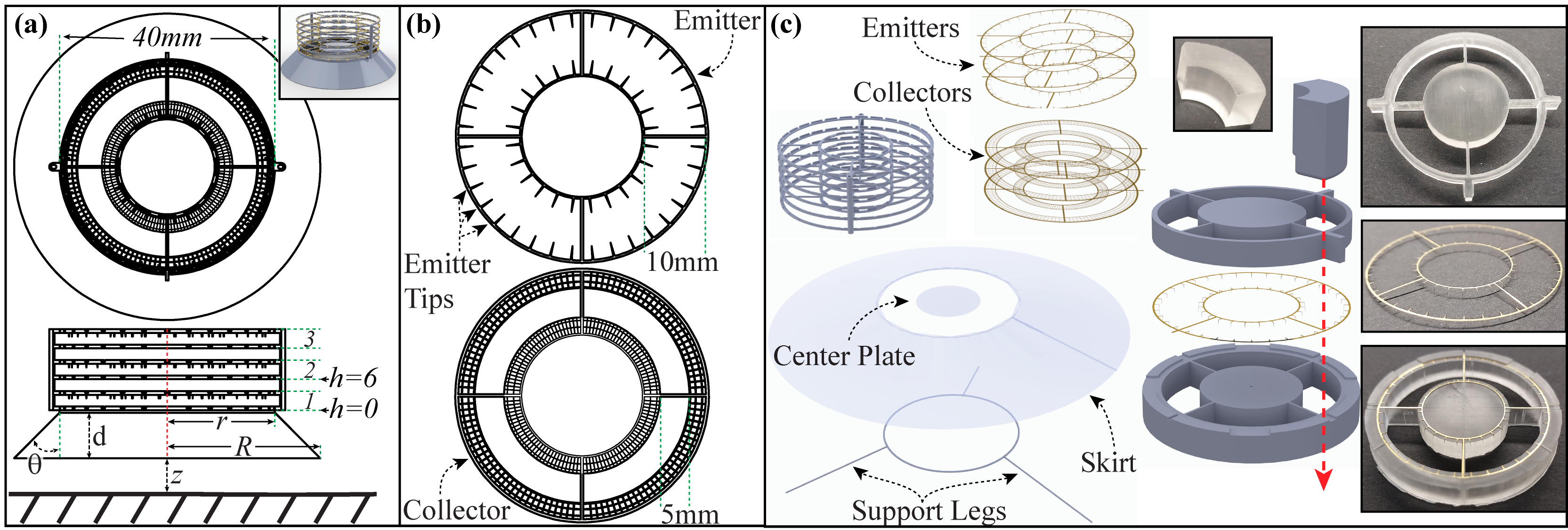}
    \vspace{-6mm}
    \caption{\textbf{(a)} Top-down and cross-section views showing parameters of interest, including the outer device diameter, number of stages ($3$), center plate position ($h$),  skirt depth ($d$), skirt angle ($\theta$), the skirt starting radius ($r$, corresponding to the thruster exhaust radius), and the exit radius ($R$). \textbf{(b)} Geometric parameters of the electrodes, with the emitter on the top and collector on the bottom. Each emitter is spaced approximately 3~mm along the circumference. \textbf{(c)} Devices are fabricated using a combination of ultraviolet-laser micromachined emitter and collector electrodes and SLA-printed structural framing, then assembled using printed jigs for both alignment and bending of the emitter tips down towards the collector. The skirt is manually cut thin PEI.}
    \label{fig:assembly}
    \vspace{-3mm}
\end{figure*}

\subsection{Thrusters and Airframe}
The propulsion core consists of three serially stacked ion acceleration stages.  Critical electrode parameters (e.g., inter-electrode gap, emitter tip spacing) are set based on results of prior empirical optimization efforts for lithographically-defined multi-stage thrusters~\cite{nelson2024high}. 

Each stage includes an emitter and collector separated by a 2~mm inter-electrode gap (Fig.~\ref{fig:assembly}). Electrodes are laser micromachined from 25~$\upmu$m brass using a 355~nm UV system, and 2~mm long emitter tips are bent out of plane at their midpoint to form vertically-oriented corona emission points using a printed stamp-and-die set (Fig.~\ref{fig:assembly}(c)). Structural components are SLA printed (Formlabs Clear V4 resin). Stages are separated by 3~mm to limit electrostatic interference while maintaining compact stacking.

Each stage employs two opposed annular emitter rows spaced 30~mm center-to-center (Fig.~\ref{fig:assembly}(a)). Emitters are spaced approximately 3~mm along the circumference of each annulus, with 10~mm internal spacing between opposing rows (Fig.~\ref{fig:assembly}(b)). The collector diameter matches the emitter footprint, and a 5~mm central aperture is removed to reduce flow obstruction without affecting corona onset.

The three-stage stack is integrated with a passive center plate and skirt (Fig.~\ref{fig:assembly}(c)). Instead of assembling a new free-standing vehicle for each configuration, we created a modular jig that enabled independent exchange of emitters, collectors, center plates, and skirts during static testing (Fig.~\ref{fig:setup}(b)). Key geometric parameters, including skirt angle $\theta$, skirt depth $d$, skirt radius $R$, and exhaust radius $r$ (Fig.~\ref{fig:assembly}(a)), were varied to quantify the impact on in-ground-effect performance.

For the free-standing vehicle, the stage stack was printed as a single structural component to reduce mass. Electrodes were inserted laterally and secured with minimal cyanoacrylate adhesive. The center plate is a 20~mm diameter disk cut from 25~$\upmu$m polyetherimide (PEI), with adjustable vertical position $h$. Skirt support legs (0.3~mm $\times$ 0.3~mm cross-section) were printed at angle $\theta$ and bonded to the 50~$\upmu$m PEI skirt to complete the assembly. 

\subsection{Experimental Setup}
\begin{figure}
    \centering
    \includegraphics[width=0.9\columnwidth]{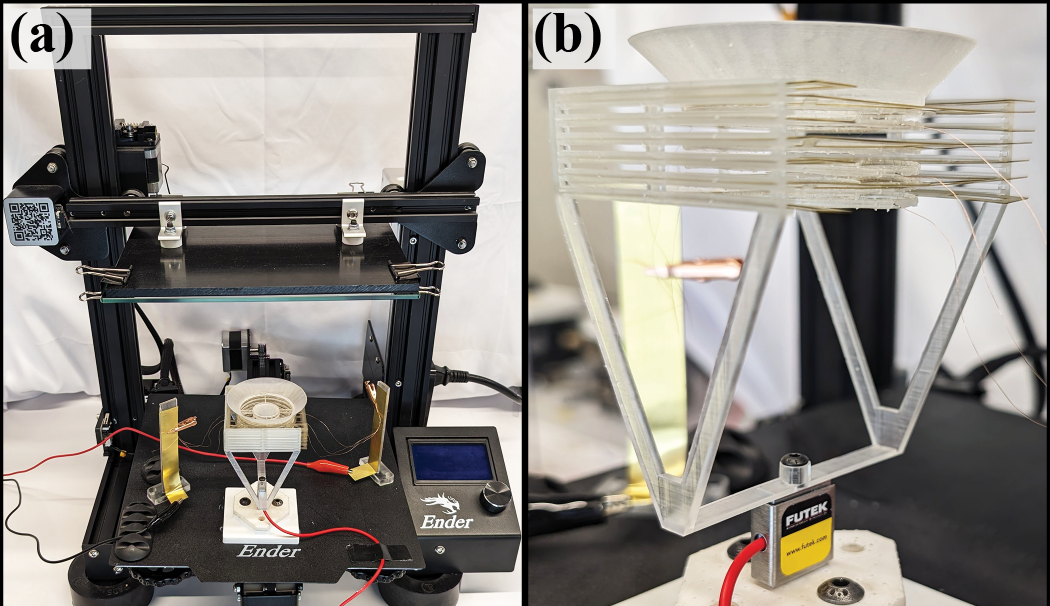}
    \vspace{-3mm}
    \caption{\textbf{(a)} The automated testing setup uses a modified 3D printer to coordinate acquisition of data with precise movement of the ground plane. The thruster is oriented with its exhaust directed upwards towards the ground plane. \textbf{(b)} Close view of the device-under-test with load cell visible. An experimental stand-in for the free-standing hovercraft enables rapid component swaps.}
    \label{fig:setup}
    \vspace{-6mm}
\end{figure}

The testbed is modeled on the ground-proximity platform validated by Nations et al. in~\cite{nations2024empirical}. A modified Ender3 Pro provides synchronized force measurement and control of ground-plane distance (Fig.~\ref{fig:setup}(a)). The extruder assembly was replaced with a 230~mm $\times$ 230~mm Delrin plate supporting a glass ground plane. Separation distance $z$ is controlled via G-code commands, enabling automated sweeps of $z/r$.

The hovercraft assembly is mounted to a FUTEK LSB200 load cell with its inlet approximately 55~mm above the bottom surface of the setup to prevent potential ceiling effects from affecting performance~\cite{nelson_empirical_2024}. Thrusters are oriented upward so that wall-jet impingement produces a measurable reaction force. High voltage is supplied by a Spellman SL8P source. Voltage and ion current are recorded via the supply monitor and shunt resistor, while load cell data are acquired at 100~ms intervals. For each geometry, voltage sweeps up to 2.5~kV are performed at fixed $z$, and in-ground-effect (IGE) performance is compared to out-of-ground-effect (OGE) measurements at large $z/r$.

\section{Results and Discussion}
\label{sec:results}
All experiments were performed with three trials per test configuration. The plotted data represent the overall mean of each configuration's trial means, with error bars indicating the standard error of the mean (SEM). In plots whose values are fractions, the error of the numerator is given.

\subsection{Baseline Ground-Proximity Performance}
\begin{figure}
    \centering
    \includegraphics[width=\columnwidth]{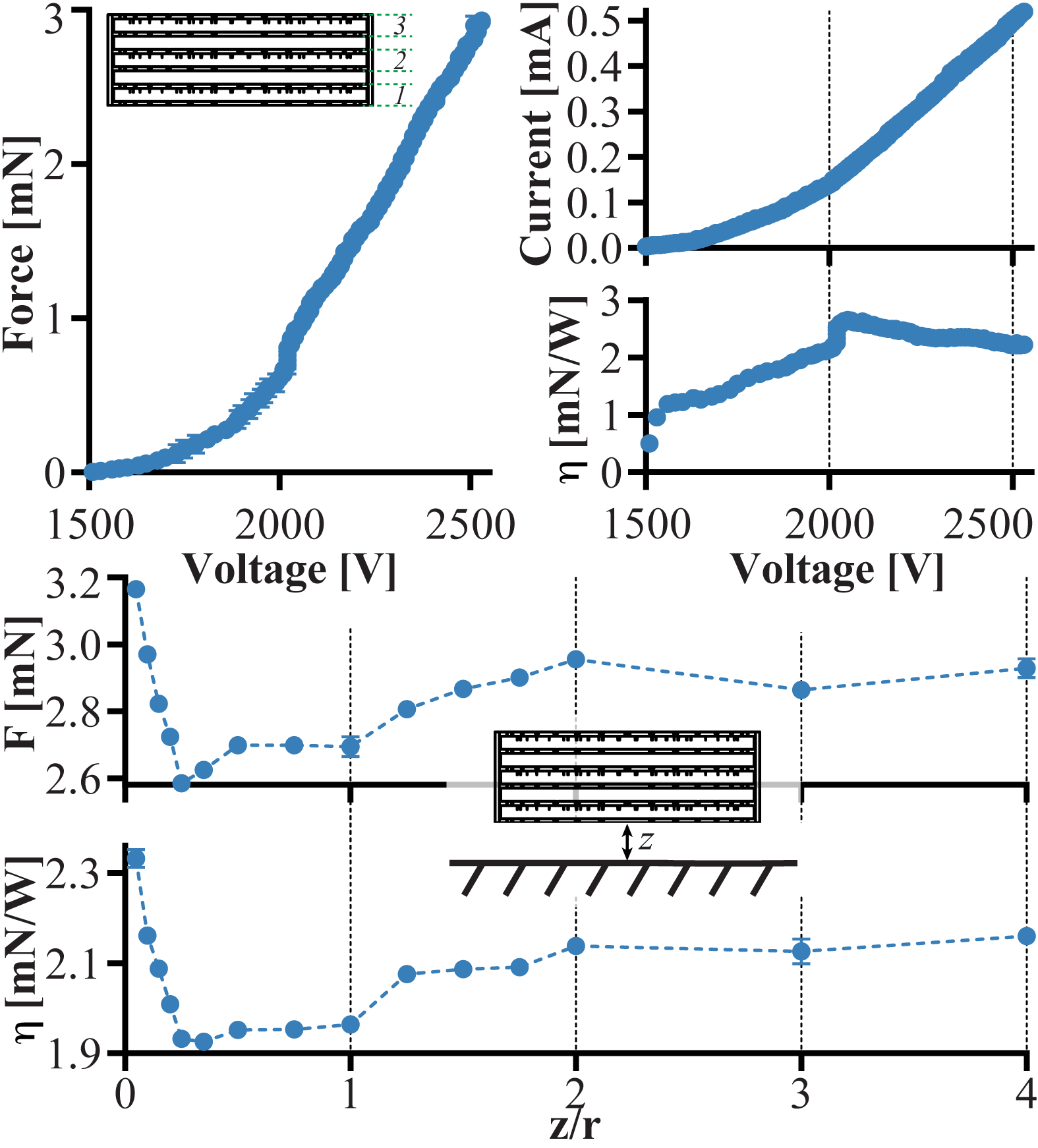}
    \vspace{-6mm}
    \caption{Results from a control device with no skirt, used as the baseline values for subsequent experiments. The force and efficiency values at a $z/r$ of 4 (bottom) serve as the out-of-ground-effect control values $F_0$ and $\eta_0$. Measurement points are dense enough to appear continuous and error is small enough to be invisible relative to marker size (top).}
    \label{fig:control}
    \vspace{-5.5mm}
\end{figure}

We first characterize the three-stage propulsion stack without a skirt or center-plate modification to establish a baseline for in-ground-effect behavior (Fig.~\ref{fig:control}). This configuration serves as the out-of-ground-effect reference, with force and efficiency denoted $F_0$ and $\eta_0$ at large $z/r$.

When operated near the ground plane, the unskirted stack exhibits only modest performance enhancement. At $z \approx 1$~mm (corresponding to $z/r \approx 0.05$), the measured thrust increases by approximately 8$\%$ relative to $F_0$, while thrust efficiency increases by approximately 4$\%$. These gains are significantly smaller than the peak proximity-induced enhancements reported for multi-thruster arrangements by Nations et al. in \cite{nations2024empirical}, which used ducted devices in contrast to the present unducted ones. The maximum achievable force is insufficient for takeoff.

These results indicate that simple exploitation of classical ground effect in a vertically impinging EAD jet is not enough to produce a pressure-supported regime for unducted actuators. Additional geometric confinement is required to meaningfully amplify proximity-induced pressure recovery.

\subsection{Skirt Angle}

\begin{figure}
    \centering
    \includegraphics[width=\columnwidth]{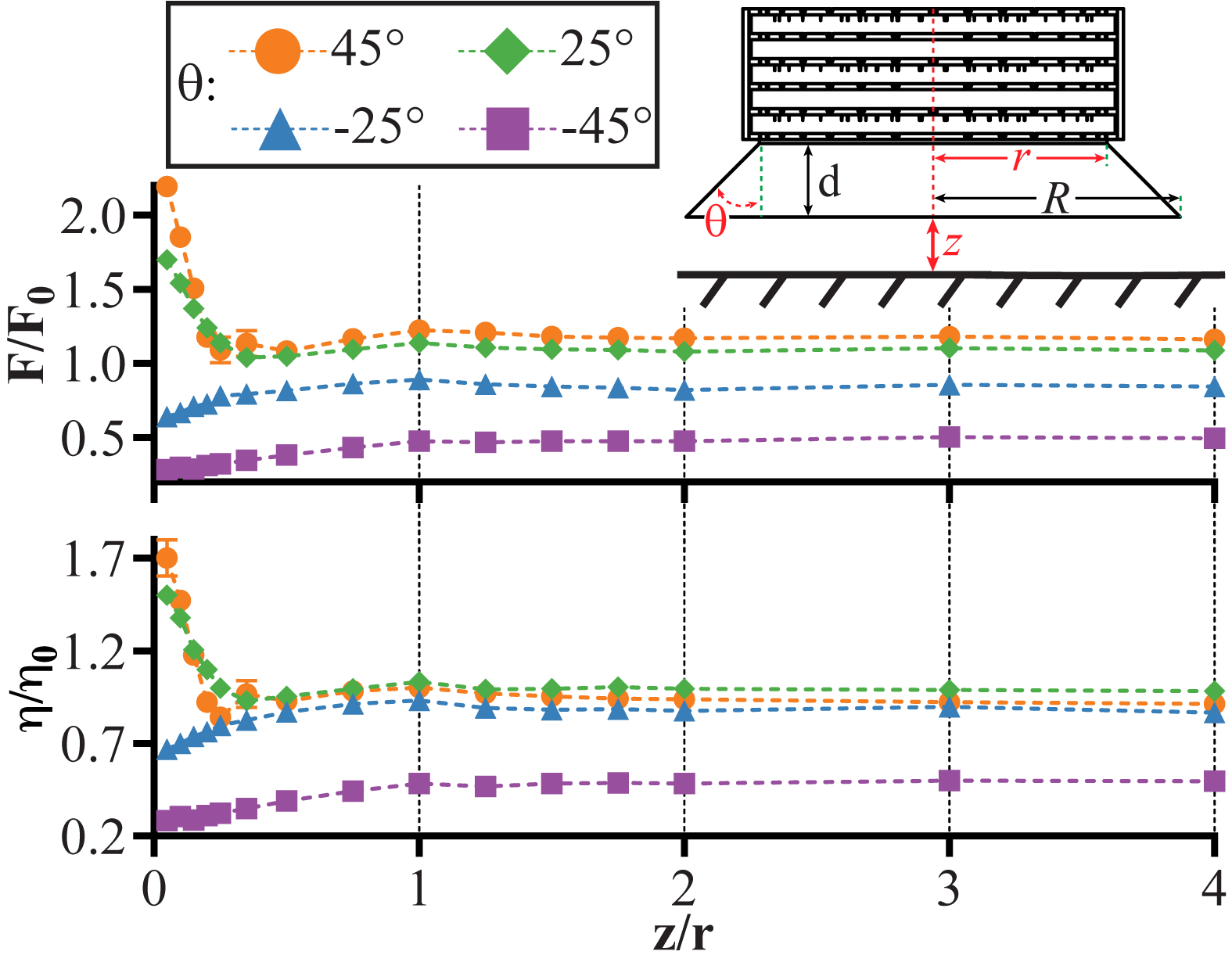}
    \vspace{-6mm}
    \caption{Force and efficiency relative to the control device's out-of-ground-effect value for different skirt angles as a function of $z/r$ ratio with both convergent and divergent skirt angles tested.}
    \label{fig:angle}
    \vspace{-6.5mm}
\end{figure}

We first compare convergent ($\theta < 0$) and divergent ($\theta > 0$) configurations at fixed skirt depth $d = 9$~mm (Fig.~\ref{fig:angle}). Convergent skirts reduce both thrust and efficiency at low $z/r$, whereas divergent skirts produce substantial gains; the 45$^{\circ}$ skirt increased force by 120\% and efficiency by 70\%. This trend suggests that divergent geometries more effectively convert the impinging jet into useful force by increasing the surface area over which elevated near-ground pressure can act. Since the net lift force scales with the pressure differential integrated over the skirt area, $F_L = \int_A \Delta p ~ dA$, increasing the area exposed to positive $\Delta p$ beneath the vehicle increases total lift without additional electrical power input. The data are consistent with this geometric pressure-recovery mechanism, as performance scales strongly with increasing $\theta$ over the tested range. 

At high $z/r$, where the skirt may be considered more of a classical exhaust nozzle, angle of convergence is shown to have a strong effect on force and efficiency, with the narrower exhaust performing the worst. This finding is consistent with prior analytical~\cite{gomez-vega_model_2023} and experimental~\cite{nelson2024high} work on multi-stage ducted (MSD) electroaerodynamic thrusters and suggests that flow is being restricted.

\subsection{Airframe Center Plate}

\begin{figure}
    \centering
    \includegraphics[width=\columnwidth]{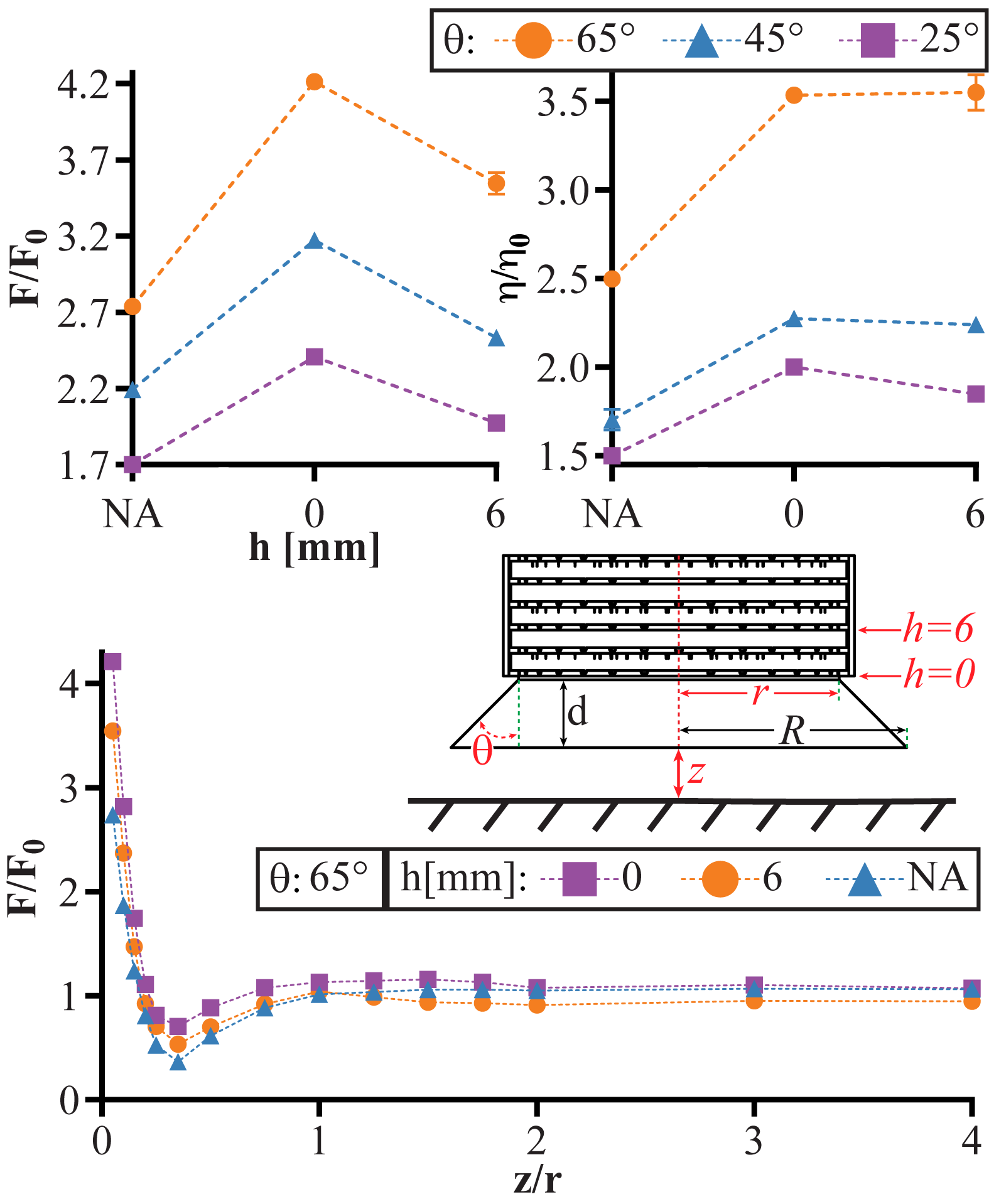}
    \vspace{-6.5mm}
    \caption{Relative force and efficiency  in-ground-effect versus center plate position and skirt angle (top) and relative force versus distance from the ground for fixed skirt angle and different center plate positions (bottom). Top data at $z = 1$~mm. $NA$ refers to configuration with no center plate.}
    \label{fig:height}
    \vspace{-5.5mm}
\end{figure}

Having established that divergent skirts substantially increase in-ground-effect lift, we next investigate the role of the center plate position $h$. The center plate modifies the underside geometry of the vehicle and therefore the distribution of pressure beneath the fuselage. It also provides an area on which fountain lift, where colliding impinged wall jets recirculate and produce a net lift force, can act.

Experiments were again conducted at fixed skirt depth $d = 9$~mm while varying skirt angle $\theta$ and center plate position $h$ (Fig.~\ref{fig:height}). The center plate was either removed, positioned flush with the base of the stage stack ($h = 0$) or elevated ($h = 6$~mm). Across all tested $\theta$, placing the center plate at $h = 0$ produced higher thrust and efficiency at low $z/r$ than the elevated configuration. A higher skirt divergence angle of 65$^\circ$ was tested here, which was shown even without a center plate to further increase force and efficiency. In all cases, addition of a center plate improved performance. The effect was most pronounced at high divergence angles; for the 65$^\circ$ skirt, a flush center plate increased relative force by over 50$\%$ and efficiency by 40$\%$ at very low $z/r$ compared to configurations with no center plate.

Physically, placing the center plate at $h = 0$ increases the effective area over which elevated near-ground pressure acts, and reduces the volume available for internal recirculation beneath the vehicle. In contrast, elevating the plate allows additional circulation beneath the propulsion core, potentially permitting fountain lift~(Fig.~\ref{fig:schematic}(b)). When the plate was elevated, force decreased while efficiency remained nearly constant relative to the $h=0$ condition; an initial hypothesis is that fountain lift is indeed present but the magnitude of internal skirt pressurization is reduced. Confirming this requires investigation with more advanced flow visualization and pressure measurement instrumentation.

A $z/r$ region dominated by aerodynamic ``suckdown'' effects, where $F/F_0 < 1$, is evident for all tested devices. This is expected behavior based on prior work~\cite{nations2024empirical}. As the hovercraft are designed for extremely low altitude flight, this may actually confer passive stability benefits. 

\subsection{Skirt Ratio}

\begin{figure}
    \centering
    \includegraphics[width=\columnwidth]{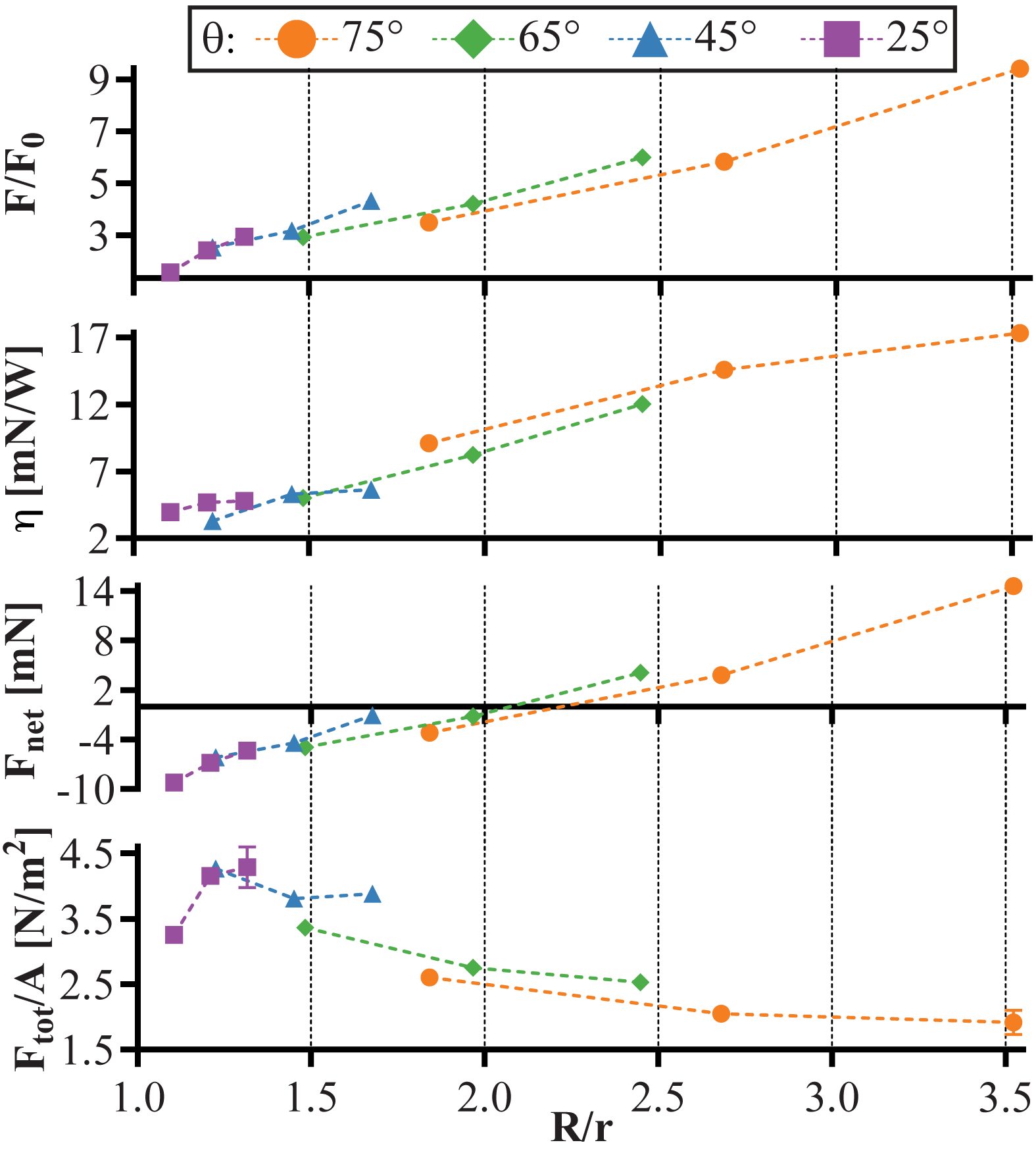}
    \vspace{-6.5mm}
    \caption{Relative force in-ground-effect, efficiency in-ground-effect, net force (produced force minus vehicle weight), and areal thrust density as functions of skirt exhaust radius $R$ to exhaust radius $r$. Skirt depth $d$ is fixed at 4.5 mm, 9 mm, and 13.5 mm for the different skirt ratios. All data were collected at approximately 2500~V at a $z$ of 1 mm (i.e., $z/r$ of 0.05).}
    \label{fig:Rr}
    \vspace{-2mm}
\end{figure}

\begin{figure}
    \centering
    \includegraphics[width=\columnwidth]{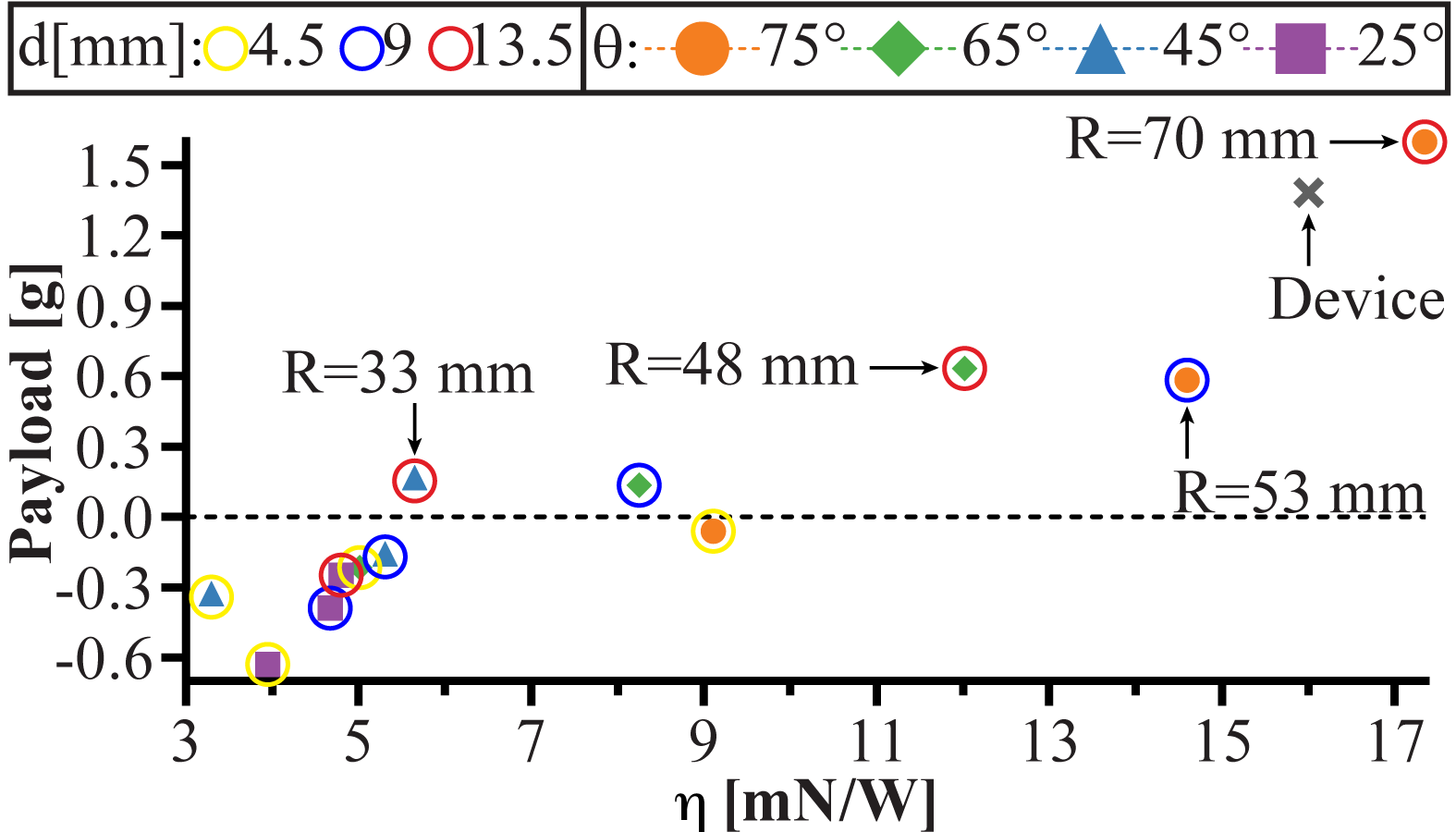}
    \vspace{-6.5mm}
    \caption{Device geometries tested during the empirical design process plotted as a function of payload and efficiency in-ground-effect. The labeled "Device" is tested data from a free-standing device assembled with the same geometry as the nearby $R$ = 70 mm point.}
    \label{fig:pareto}
    \vspace{-5.5mm}
\end{figure}

To further increase pressure-supported lift, we varied skirt depth $d$ while holding the center plate at $h = 0$. Increasing $d$ at a fixed angle increases the skirt exhaust radius $R$, and thus the skirt ratio $R/r$, where $r$ is the thruster exhaust radius (Fig.~\ref{fig:Rr}). This parameter captures the effective area over which elevated near-ground pressure can act.

Experiments were conducted for $\theta \in {25^\circ, 45^\circ, 65^\circ, 75^\circ}$ and $d \in {4.5, 9, 13.5}$~mm, corresponding to increasing $R/r$. At fixed $\theta$, thrust in-ground-effect increases monotonically with $R/r$ over the tested range. The largest measured thrust enhancement occurs for $\theta = 75^\circ$ and the maximum tested $R/r$, yielding approximately an 840$\%$ increase relative to the unskirted baseline at $z \approx 1$~mm.

Net lift, defined as the marginal lift contributed by the skirt at $z \approx 1$~mm, was computed as $F_{net} = F_{tot} - F_{0} - W_{vehicle}$, where $W_{vehicle}$ includes the mass of the stage stack, skirt, and structural components, $F_{tot}$ refers to the measured force at $z \approx 1$~mm, and $F_{0}$ refers to force produced by the baseline configuration at $z \approx 1$~mm. Several configurations achieved $F_{net} > 0$ at $z \approx 1$~mm, indicating a thrust-to-weight ratio greater than unity in-ground-effect.

Figure~\ref{fig:pareto} maps payload (derived from net force) against efficiency for all tested skirt angles $\theta$ and skirt ratios $R/r$. Increasing $R/r$ increases payload monotonically by expanding the pressure-recovery area. The largest tested geometry ($\theta = 75^\circ$, $R=$70 mm) achieves the highest payload and efficiency, at the lowest areal thrust density. Devices with a skirt exhaust radius $R$ below approximately 30~mm exhibit insufficient thrust-to-weight ratio for lift-off. The feasible hover region therefore lies along a design frontier balancing compactness against lift capability. While not explored directly in this work, skirt volume also will have an effect on vehicle stability and control authority. 

\begin{figure*}[t]
    \centering
    \includegraphics[width=\textwidth]{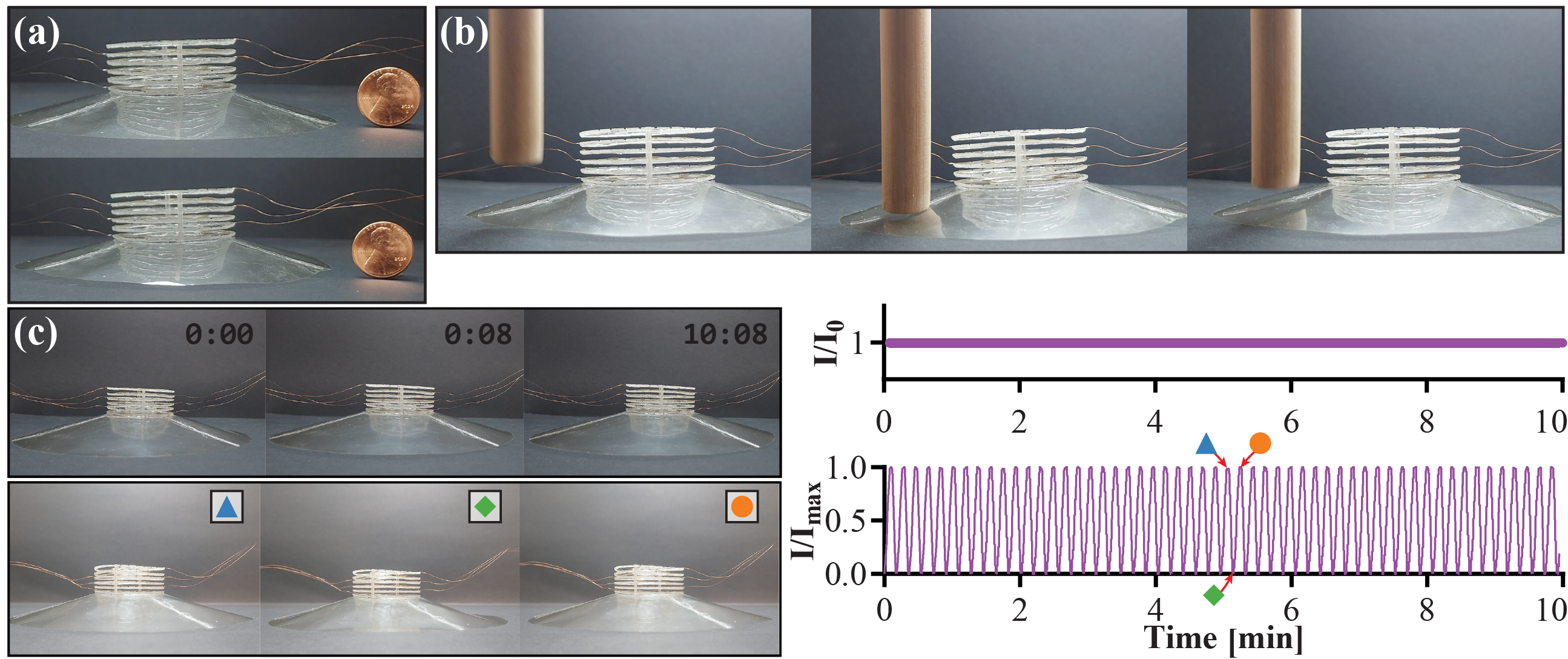}
    \vspace{-6mm}
    \caption{\textbf{(a)} The hovercraft takes flight at an applied voltage of approximately 2500~V.  \textbf{(b)} After being struck with a stick in the thruster core or the skirt, the hovercraft returns to its equilibrium height and stabilizes its attitude passively.  \textbf{(c)} The hovercraft can sustain tethered flight for over ten minutes (top) and complete high numbers of takeoff and landing cycles (bottom) without  observable change in ion current, $I$, where $I_0$ refers to the measured ion current directly after takeoff and $I_{max}$ refers to the maximum measured ion current during the first takeoff cycle.}
    \label{fig:lifetimes}
    \vspace{-2mm}
\end{figure*}

\subsection{Lift-Off and Hover Demonstration}

The vehicle achieves stable hover at approximately 1–2~mm above the ground plane without active attitude control (Fig.~\ref{fig:lifetimes}(a)). The equilibrium height emerges passively from the balance between pressure-supported lift and increased flow losses at larger separation distances. When perturbed laterally or vertically, the hovercraft exhibits restoring behavior consistent with a proximity-induced stiffness effect (Fig.~\ref{fig:lifetimes}(b)). Small increases in height reduce lift, while decreases in height increase stagnation pressure beneath the vehicle. This passive feedback produces a stable hover equilibrium without closed-loop control. 

Extended flight tests showcase the durability of the design (Fig.~\ref{fig:lifetimes}(c)). Flight times of over 10 minutes were recorded, with no observed change in ion current (a proxy for force) over the duration. This implies that the corona discharge electrodes are not experiencing corrosion, oxidation, or kinetic pitting at the operating point necessary for flight. Repeated takeoff-and-landing cycles (54 over a 10 minute period) also had no measurable effect on ion current, implying plasma ignition is not harmful to the device. Importantly, this shows that through a focus on fabrication and assembly consistency, it is possible to eliminate destructive arcing in large-scale multi-stage EAD devices with sub-centimeter electrode gaps.

\subsection{Free-standing Hovercraft Characterization}

\begin{figure}[t]
    \centering
    \includegraphics[width=0.9\columnwidth]{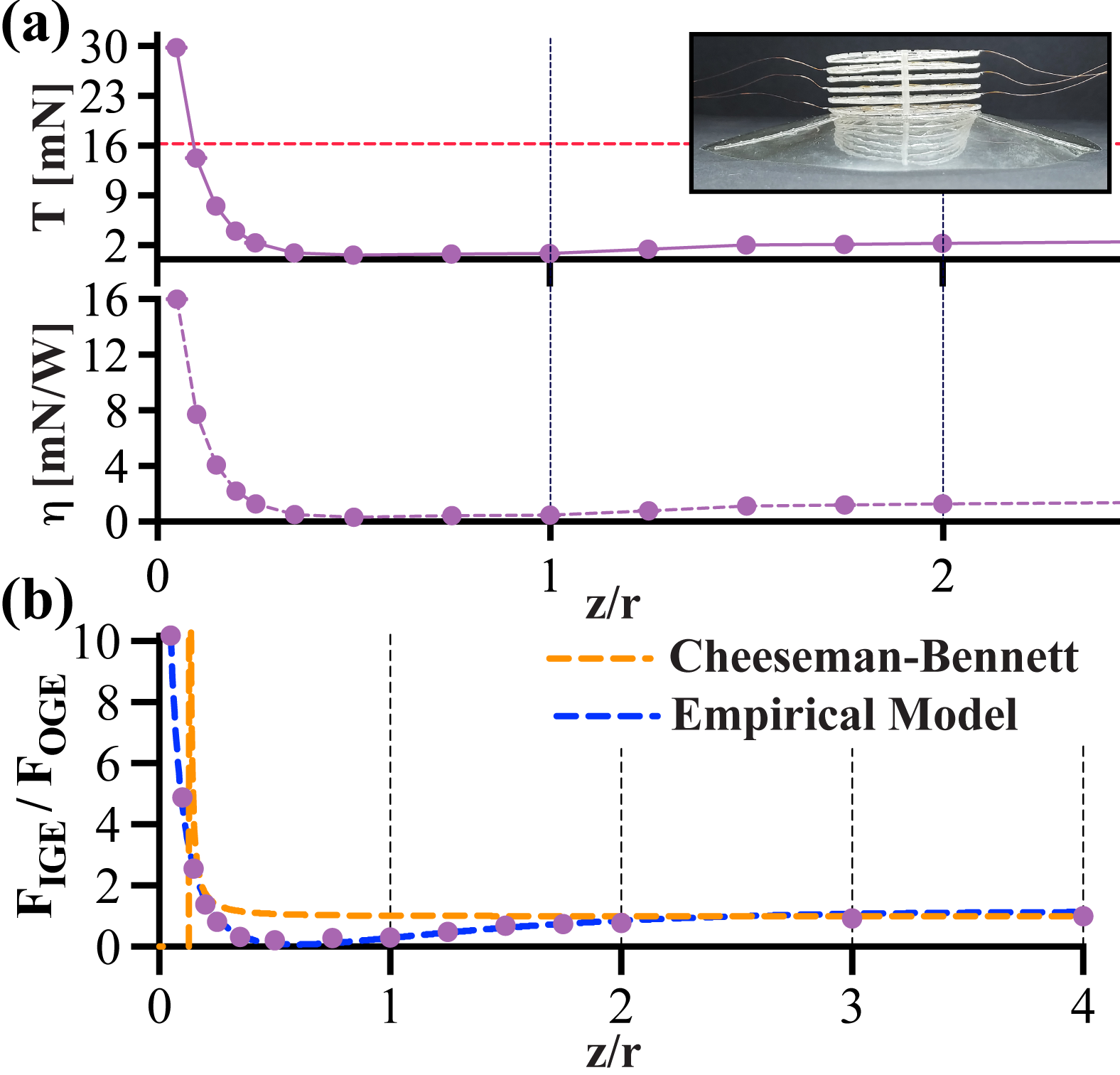}
    \vspace{-3.5mm}
    \caption{\textbf{(a)} Thrust and thrust efficiency versus $z/r$ for a device with a skirt angle of $75^{\circ}$, $d$ of 13.5 mm, and $h$ of 0 mm. The red dashed line represents the total measured weight of the device. \textbf{(b)} Cheeseman-Bennett (Eq. 1) and proposed empirical model (Eq. 2) fits to the data. Cheeseman-Bennett $k$ = 0.1309; Empirical Model $A=56.18$, $B=4.263$, $k=0.01547$.}
    \label{fig:selected}
    \vspace{-6.5mm}
\end{figure}

\begin{figure}[t]
    \centering
    \includegraphics[width=\linewidth]{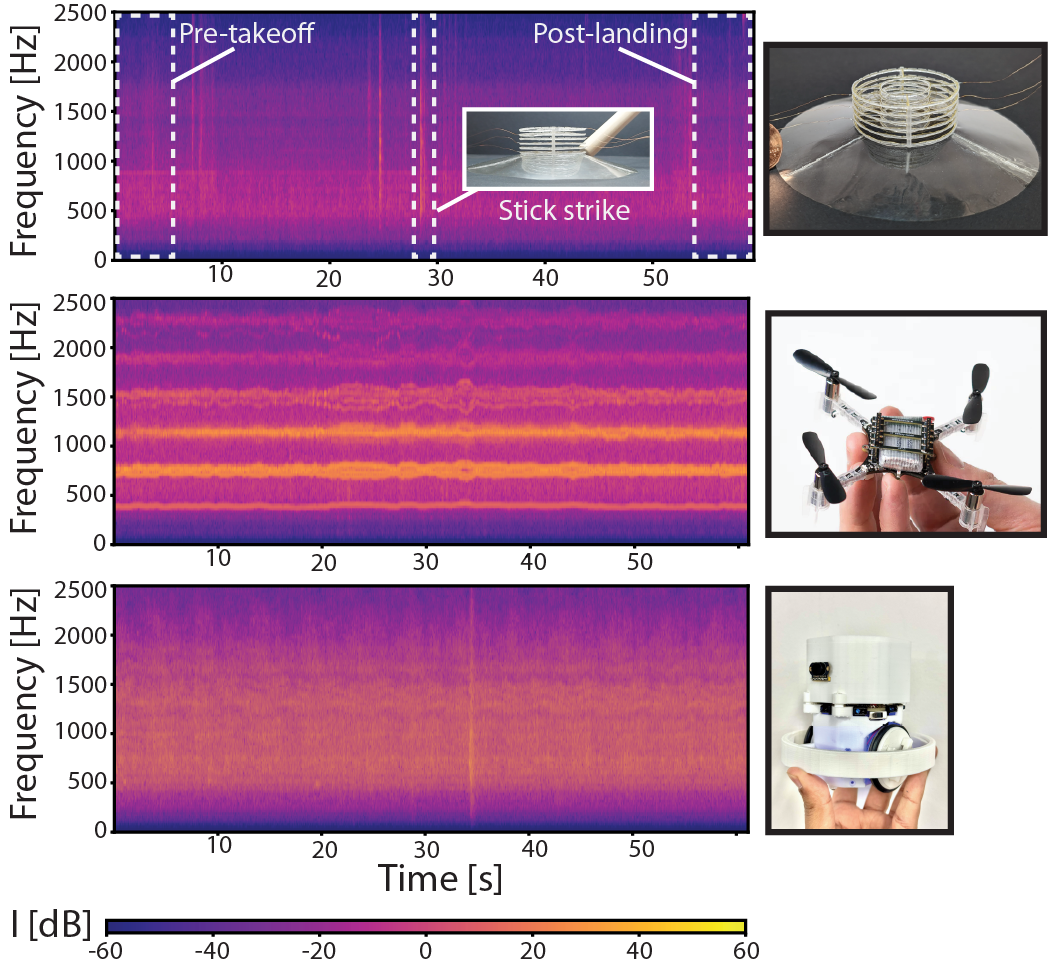}
    \vspace{-6.5mm}
    \caption{Audio extracted from a video of the hovercraft in flight shows that it is virtually silent. Audio from videos of a similarly sized quadrotor (Crazyflie 2.1) and differential drive robot (HeRo 2.0) at approximately the same distance show that they produce highly tonal and broadband noise, respectively, at much higher intensity.}
    \label{fig:audio}
    \vspace{-2mm}
\end{figure}

\begin{figure} [h]
    \centering
    \includegraphics[width=0.9\linewidth]{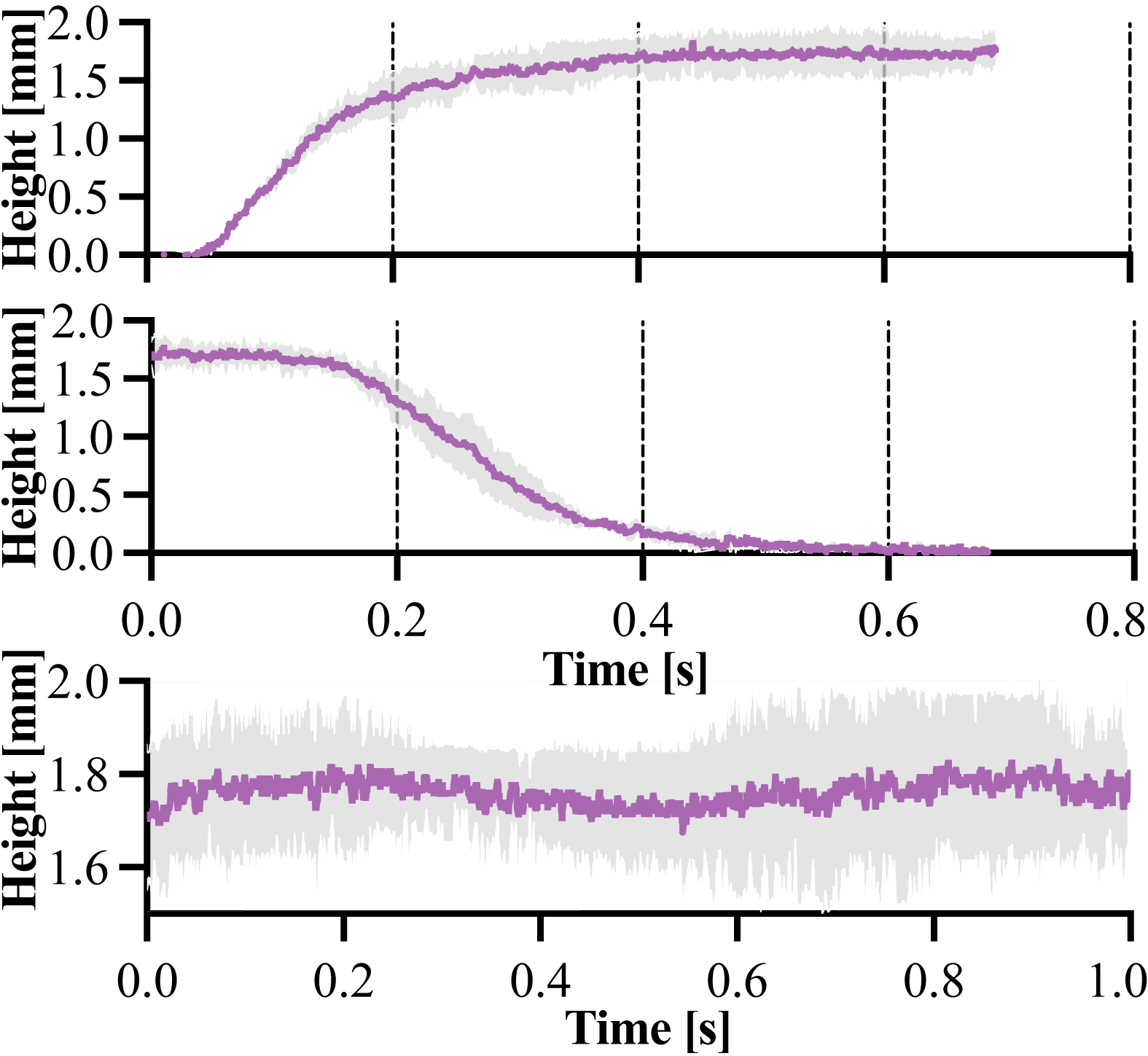}
    \vspace{-3.5mm}
    \caption{Height versus time during takeoff and landing (top) and stable hovering (bottom), extracted from high-speed video. Purple curve shows the mean height across trials, and the gray shaded region is standard deviation.}
    \label{fig:takeoff_landing}
    \vspace{-6mm}
\end{figure}

A free-standing hovercraft with a skirt angle $\theta$ of 75$^{\circ}$ and $d=13.5~\text{mm}$ was also tested on the static measurement setup. Figure~\ref{fig:selected}(a) shows the approximately 1.6 gram hovercraft achieves a maximum thrust of almost 30 mN (a thrust-to-weight ratio of approximately 1.8) at an efficiency of 16 mN/W, an order-of-magnitude improvement over metrics previously reported for EAD-propelled MAVs capable of lift-off (Table~1)~\cite{drew2018toward}.

\begin{table}[t]
\centering
\caption{Comparison of MAV-scale EAD-propelled flying robots.}
\vspace{-3mm}
\label{tab:ead_mav_comparison}
\setlength{\tabcolsep}{2.2pt}
\renewcommand{\arraystretch}{1.08}
\begin{tabularx}{\columnwidth}{@{}p{0.24\columnwidth}p{0.20\columnwidth}p{0.23\columnwidth}p{0.25\columnwidth}@{}}
\toprule
Paper & Characteristic Length [cm] & Thrust Efficiency [mN W$^{-1}$] & Payload Capacity [mg] \\
\midrule
Drew et al.~\cite{drew2018toward}
& $2.0$
& $2.0$
& $70$ \\

Prasad et al.~\cite{prasad_laser-microfabricated_2020}
& $2.5$
& $3.3$
& $75$ \\

Zhang et al.~\cite{zhang_centimeter-scale_2022}
& $6.5$
& $2.95$
& $137$ \\

Gu et al.~\cite{gu_microrobotic_2024}
& $4$
& $< 1$
& $72$ \\

\textbf{This Work}
& $14$
& $16$ 
& $1490$ \\

\bottomrule
\end{tabularx}
\vspace{-6.5mm}
\end{table}

The Cheeseman-Bennett ground effect model with dimensionless fit parameter $k$ is  widely used to characterize ground effect vehicles despite known limitations~\cite{cheeseman_effect_1955,conyers_empirical_2018}:
\vspace{-1mm}
\begin{equation}
\vspace{-2mm}
\frac{F_{\mathrm{IGE}}}{F_{\mathrm{OGE}}} = \frac{1}{1 - (\frac{k}{z/r})^2}
\end{equation}

We fit this model to our data ($k=0.1309$) to test the hypothesis that it is unsuitable for EAD-propelled hovercraft, confirming that the model fails to capture both the suckdown effect at moderate $z/r$ and the shape of the low $z/r$ ground effect thrust magnification, in part due to the model's discontinuity when $k = z/r$ (Fig.~\ref{fig:selected}(b)). We propose an alternative empirically derived model that captures both in-ground-effect thrust enhancement and suckdown behavior:
\begin{equation}
\frac{F_{\mathrm{IGE}}}{F_{\mathrm{OGE}}} = 1 + \frac{A}{1 + \frac{z/r}{k}} - B e^{-z/r},
\end{equation}
where $A$ is correlated with the ground-effect thrust enhancement and $B$ with the suckdown contribution, and $k$ controls the decay rate with normalized height. Fit parameters of $A=56.18$, $B=4.263$, and $k=0.01547$ are shown to fit device data (Fig.~\ref{fig:selected}(b)).

\subsection{Acoustic Signature}

The audio track from the disturbance rejection video taken for Fig.~\ref{fig:lifetimes}(b) was analyzed and compared to video taken at comparable distances of a Bitcraze Crazyflie 2.1 micro air vehicle and a HeRo 2.0~\cite{rezeck2023hero} differential-drive swarm robot (Fig.~\ref{fig:audio}). The ion-propelled hovercraft is virtually silent, with the spectra dominated by the ambient (primarily HVAC) room noise. This contrasts with the Crazyflie and HeRo, which produce loud tonal and broadband noise, respectively. 

\subsection{Vehicle Dynamics}

A high-speed camera (Kron Tech. Chronos 1.4) was used to collect 1069~fps and 3030~fps video of the free-standing hovercraft in front of a 2~mm calibration grid. Height data were extracted from the videos using a Python tracking script, where the hovercraft contour was detected within a selected region of interest and the topmost contour point was used to estimate vertical position in each frame (Fig.~\ref{fig:takeoff_landing}). Across multiple trials the average takeoff velocity was calculated as 9.6~$\pm$~1~mm/s, and average landing velocity was -7.3~$\pm$~0.5~mm/s, with an equilibrium hover height of about 1.75~$\pm$~0.1~mm. Unlike for a typical VTOL-capable flier, a damped response with stable altitude is evident despite open-loop control. Further exploration of these dynamics is warranted to quantify relevant drag forces (e.g., squeeze-film damping) in this low altitude and Reynolds number regime. 

\section{Limitations and Future Work}
This study does not exhaustively explore the propulsion design space. The annular electrode diameter, inter-electrode gap, and inter-stage spacing were held fixed during optimization based on prior work and preliminary testing, but multi-parameter optimization could further improve efficiency and payload capacity. Such optimization will require more repeatable fabrication methods, since reliable assembly of the free-standing vehicle remains sensitive to electrode geometry and manual alignment.

Reynolds number effects and near-ground pressure fields were also not directly measured. An ideal momentum-flux estimate from out-of-ground-effect thrust gives an effective jet Reynolds number of approximately $3.8 \times10^3$, but direct flow and pressure measurements will be needed to predict scaling across vehicle sizes, operating conditions, and skirt geometries, and enable more direct comparison to other MAV architectures.

The current prototype also relies on an external high-voltage supply. Autonomy will require integration of a lightweight high-voltage DC–DC converter~\cite{park_lightweight_2020}, together with electrode segmentation for differential thrust generation and closed-loop control. Future work will combine motion capture, high-speed video, and system identification to refine the vehicle dynamics model and control strategy. Finally, because the hovercraft operates very close to the ground, its performance may also depend strongly on nearby obstacles and surface nonuniformity; these effects must be characterized before controlled flight in realistic confined environments can be achieved.

\section{Conclusion}
We presented an ion-propelled hovercraft capable of silent and sustained hover, repeated flight cycling, and disturbance rejection, while in close proximity to a ground surface. Building upon prior observations of proximity-induced performance enhancement, we identified and explored key geometric parameters, most notably skirt angle $\theta$ and skirt ratio $R/r$, that amplify pressure-supported lift and enable thrust-to-weight ratio greater than unity. The optimized configuration (skirt $\theta = 75^\circ$, $d = 13.5$~mm) achieved a maximum efficiency of 16~mN/W and a maximum thrust of 29.81~mN in-ground-effect, the latter representing an over 800$\%$ improvement relative to the baseline configuration. While the ideal skirt geometry is thruster specific, qualitative insights from this investigation are expected to hold within broadly similar size scales and thrust magnitudes. Ultimately, this work establishes a scalable design framework for pressure-supported EAD propulsion and provides a foundation for future development of near-silent micro hovercraft.





\bibliographystyle{IEEEtran}

\bibliography{prox}

\end{document}